\documentclass{article}
\usepackage{spconf,amsmath,graphicx, tabularx}

\title{Optimizing Two-Pass Cross-Lingual Transfer Learning:  Phoneme Recognition and Phoneme to Grapheme Translation}
\name{Wonjun Lee $^1$, Gary Geunbae Lee $^{1,2}$, Yunsu Kim $^3$}

\address{
	 $^1$ Department of Computer Science and Engineering, POSTECH, Republic of Korea \\
	 $^2$ Graduate School of Artificial Intelligence, POSTECH, Republic of Korea \\
    $^3$ aiXplain Inc. Los Gatos, CA, USA \\
	\{lee1jun, gblee\}@postech.ac.kr, yunsu.kim@aixplain.com}

\copyrightnotice{979-8-3503-0689-7/23/\$31.00~\copyright2023 IEEE}
\begin{document}
%
\maketitle

\begin{abstract}
This research optimizes two-pass cross-lingual transfer learning in low-resource languages by enhancing phoneme recognition and phoneme-to-grapheme translation models. Our approach optimizes these two stages to improve speech recognition across languages. We optimize phoneme vocabulary coverage by merging phonemes based on shared articulatory characteristics, thus improving recognition accuracy. Additionally, we introduce a global phoneme noise generator for realistic ASR noise during phoneme-to-grapheme training to reduce error propagation. Experiments on the CommonVoice 12.0 dataset show significant reductions in Word Error Rate (WER) for low-resource languages, highlighting the effectiveness of our approach. This research contributes to the advancements of two-pass ASR systems in low-resource languages, offering the potential for improved cross-lingual transfer learning.

\end{abstract}
\section{Introduction}
\label{sec:intro}

Automatic Speech Recognition (ASR) is essential in various applications, but the reliance on large labeled datasets poses challenges for low-resource languages. In these scenarios, end-to-end (E2E) systems may face limitations. We propose a phonetic-level approach that enhances phoneme recognition and phoneme-to-grapheme (P2G) translation models to address these challenges. Our two-pass ASR system first performs phoneme recognition and then translates the recognized phonemes into graphemes, allowing for improved performance in low-resource languages.


When choosing modeling units for ASR, the decision of- ten lies between grapheme \cite{JUST, M-CTC, multilingual} and phoneme sets \cite{Simple, UniSpeech, Globalphone}. Using grapheme units in multilingual ASR encounters challenges due to the lack of common graphemes across languages. In contrast, phoneme units facilitate the learning of shared phonetic representations, making cross-lingual transfer learning effective. The International Phonetic Alphabet (IPA) \cite{IPA} is well-suited for cross-lingual phoneme representation.


Rather than relying on grapheme units, which may not leverage the advantages of cross-lingual transfer learning, we choose to utilize phoneme units as the final output of our ASR system. However, directly generating transcriptions using phoneme-based models presents challenges, requiring additional steps such as fine-tuning the output layers with grapheme units or employing additional neural network models to convert phoneme output into grapheme.


To improve the accuracy of ASR, we incorporate a dedicated translation model that converts phoneme outputs into grapheme units. This approach offers several advantages, especially for low-resource languages with limited labeled speech data. By leveraging a substantial amount of text data, we can enhance the performance and applicability of our approach in such settings.

However, this two-pass approach may still encounter potential challenges. The IPA used for cross-lingual phoneme representation may not efficiently capture the shared phonetic characteristics across languages. Moreover, phoneme recognition results can be inaccurate, leading to error propagation during the phoneme-to-grapheme (P2G) translation step.

To address these challenges and optimize our two-pass ASR system, we focus on two key aspects: enhancing phoneme vocabulary coverage and mitigating error propagation. In this regard, we propose a novel approach called Pivot Phoneme Merging (PPM), which groups phonemes based on shared articulatory features. By facilitating improved vocabulary sharing across languages, our phoneme merging approach enhances the overall performance and effectiveness of the ASR system, particularly in scenarios where there are significant variations in phoneme vocabulary among different languages.

In addition, we present a Global Phoneme Noise (GPN) generator. This noise generator enables the pseudo-labeling of external text corpora, incorporating realistic ASR noise into the training process for P2G translation. By leveraging this noise generator, we can effectively reduce error propagation and enhance the overall robustness of the P2G translation process. Our methodology aims to advance ASR systems' effectiveness in low-resource languages. The following sections delve into the experimental evaluation and results, demonstrating the potential of our approach for improving cross-lingual transfer learning in low-resource scenarios.

\section{Related works}
\label{sec:format}

\subsection{Phoneme Recognition}

Several studies (\cite{JUST, multilingual, Simple, UniSpeech, survey}) have explored the challenges of cross-lingual transfer learning in low-resource languages. One approach introduced by \cite{Simple} utilized wav2vec2.0 \cite{wav2vec2} and combined various language datasets using a single combined IPA vocabulary in over 50 languages. To handle out-of-vocabulary (OOV) tokens, they employed a mapping function based on articulatory/phonological features, connecting phonemes between training and target vocabularies.

It is essential to mention that the previous study \cite{Simple} did not take into account the coverage of vocabulary across languages during the training process. Instead, they relied on the mapping function to address OOV tokens without specifically optimizing vocabulary coverage. In contrast, \cite{allosaurus} proposed a shared and language-specific phone set-based approach to tackle the vocabulary coverage problem.

In our proposed approach, we build upon the work of \cite{Simple} by emphasizing optimizing vocabulary coverage. By combining only 10 languages (compared to 50+ in \cite{Simple}) into a single combined phoneme vocabulary, our PPM method  efficiently merges similar phonemes, maximizing the coverage of phoneme representations across languages. This approach enables more effective cross-lingual transfer learning, particularly in low-resource language scenarios.

\subsection{Two-Pass Automatic Speech Recognition}

The concept of translation-based Two-Pass ASR has gained attention in the research literature as a powerful approach for improving ASR system accuracy and robustness. This framework consists of two main stages: signal-to-text and text-to-text processing. The first stage, signal-to-text processing or acoustic modeling, focuses on converting the audio input (speech signals) into textual representations. This stage involves capturing the acoustic features of the speech signals and mapping them to corresponding text units such as phonemes or graphemes. The second stage, the text-to-text processing or the translation model, is responsible for further refining the textual representation obtained from the acoustic model. This step can involve correcting spelling mistakes, enhancing grammar, or converting intermediate texture representation into target representation. The Two-Pass ASR framework combines signal-to-text and text-to-text processing stages to provide a comprehensive approach that can improve the performance of ASR systems.

Prior studies \cite{sc-BERT, sc-kfold, sc-rnnt, sc-TTS} have focused on the idea of a translation model that corrects noisy ASR hypotheses, known as spelling correction. These studies aimed to convert inaccurate ASR outputs into clean and correct transcriptions. However, these studies were primarily designed for a monolingual setup, and their main goal was to perform spelling correction rather than involving P2G translation.
In the field of two-pass ASR with P2G translation, a notable study by \cite{first} focuses on utilizing phonemes as an intermediate representation. They introduce a comprehensive two-pass ASR system incorporating phoneme recognition and P2G translation stages. Similarly, \cite{tranUSR} investigated using a two-pass ASR system for cross-lingual transfer learning. Their study demonstrated the ef- effectiveness of incorporating phoneme recognition as the initial pass, followed by the translation of recognized phonemes into words (P2W). This approach offers improved handling of language-specific variations and enhances overall ASR system performance. Even though \cite{tranUSR} introduced a multilingual P2G approach, the investigation of P2G in a multilingual setup is still limited. While phonemes serve as intermediate recognition units, our ultimate goal is to generate accurate grapheme outputs, which may require modifications or simplifications to improve phoneme recognition and translation results.

Additionally, P2G translation can be further enhanced through training with noisy text data, enabling robust performance in noisy ASR hypotheses. Previous studies such as \cite{sc-BERT, sc-kfold, first} have employed the K-fold method to generate ASR noise for training the translation model. Another approach, as seen in  \cite{sc-TTS}, involves generating synthetic audio and applying ASR inference to produce noisy data for a translator. In our research, we propose a novel solution called the Global Phoneme Noise (GPN) generator, which generates realistic ASR noise directly from plain text corpora.

By introducing the GPN, we enhance the training process for P2G translation by incorporating realistic noise into the training data. This noise generation technique aids in capturing the challenges and variations present in real-world ASR scenarios, ultimately improving the robustness and accuracy of the P2G translation process. Our research expands the capabilities of P2G translation by integrating noise-aware training strategies, resulting in more effective and reliable cross-lingual transfer learning in ASR systems.

\begin{figure*}[t]
    \centering
    \includegraphics[width=0.80\textwidth]{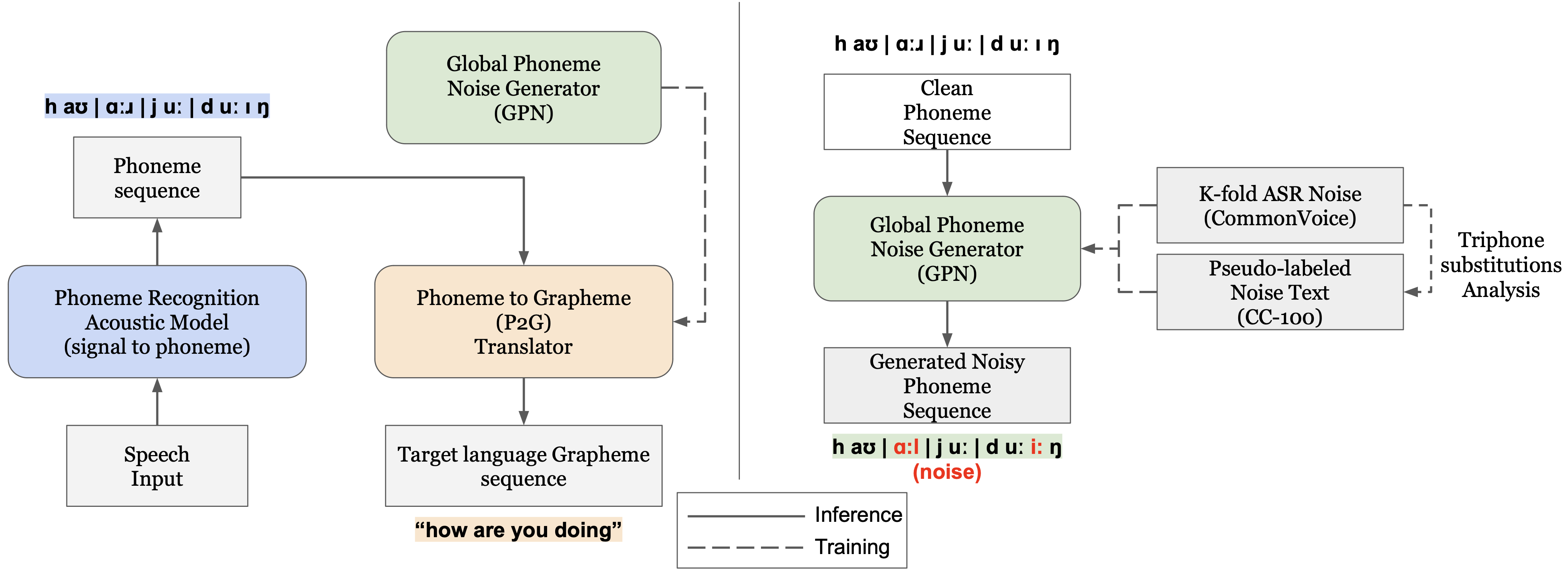}
    \caption{Overall architecture of Two-Pass ASR system with Phoneme Recognition and Phoneme-to-Grapheme Translator (Left) and Global Phoneme Noise (GPN) generator training process (Right)}
    \label{fig:arch}
\end{figure*}

\section{Method}
\label{sec:pagestyle}

We developed two methodologies, Pivot Phoneme Merging (PPM) and Global Phoneme Noise (GPN) generator, to improve cross-lingual transfer learning in low-resource languages. Our approach involves a two-pass ASR system comprising two main stages: phoneme recognition and phoneme-to-grapheme (P2G) translation (Figure \ref{fig:arch}). PPM optimizes phoneme vocabulary coverage by merging phonemes based on shared articulatory characteristics. At the same time, GPN addresses the limited availability of training data by incorporating pseudo-labeled noise into clean text.

\subsection{Pivot Phoneme Merging (PPM)}
\label{sec:method:ppm}
The PPM (Pivot Phoneme Merging) enables cross-lingual learning by creating a shared vocabulary that maximizes sharing among multiple languages. To construct this shared vocabulary, we calculate the occurrence frequencies of phoneme sets in each language, determining the probability of their occurrence. Based on these probabilities, we compute importance scores for each phoneme by summing up their scores across all languages. The top $K$ pivots with the highest scores are selected as reference points for merging non-pivot phonemes in each language.

To measure the similarity between pivot phonemes and other phonemes, we calculate the phonetic distance by utilizing the Panphon tool \cite{panphon}, which incorporates articulatory feature vectors. The humming feature edit distance, ranging from 0 to 3.0, is used in our experiments, with values closer to 0 indicating higher similarity.

For instance, when selecting 30 pivots with a threshold value 2.0 (K=30, T=2.0), each language matches non-pivot phonemes to the chosen pivots. If the humming edit distance exceeds 2.0 ($T$), the phonemes are kept as individual tokens rather than merged. 

This approach allows us to capture frequently occurring and important tokens across languages while preserving phonemes that may be less common but hold significance in specific languages.






\subsection{ASR noise for P2G translation}
\label{sec:noise}

The performance of the translation model, which corresponds to the second pass in the two-pass ASR model, significantly impacts the final ASR results by effectively handling the noise generated in the first step (phoneme recognition).
Previous studies on spelling correction, which also utilizes the two-pass approach \cite{ sc-BERT, sc-kfold, sc-rnnt}, obtained ASR noise text data through K-fold training and used it for training in the second step. However, the available training data is minimal for low-resource languages, making it insufficient to obtain adequate ASR noise data using only the K-fold method.
We propose a pseudo-labeling method using K-fold-generated noise on plain text data to address this problem and improve the robustness of phoneme-to-grapheme (P2G) translation.

\textbf{K-fold}: We employed the K-fold method to generate noisy ASR transcriptions. The train set was divided into 10 folds, and we trained 10 separate models. Each model utilized 9 folds for training, while the remaining fold was used for inference. This approach enabled us to obtain noisy ASR transcriptions for the entire train set, thereby enhancing the robustness of the P2G training process.

\textbf{Triphone Pseudo-Noise Labeling:} To address the limited availability of training data for specific languages, we conducted triphone noise analysis on the noisy ASR transcriptions obtained through the K-fold method. This analysis allowed us to identify patterns of phoneme substitutions and consider the triphone context by examining three consecutive phonemes.

Based on the analysis results, we paired the substituted triphones with their corresponding clean (original) triphones. We calculated the frequency of each triphone ($freq$) and the frequency of its correct form in the clean original data ($total\ freq$).

For pseudo-noise labeling, we formulated a replacement probability using the following equation:
\footnotesize $$ \text{Replacement Probability} = \left(\frac{{\text{freq}}}{{\text{total freq}}}\right) \times \left(\frac{{3.0 - \text{distance}}}{3.0}\right) $$ \normalsize

The $distance$ parameter represents the humming edit distance (see Section \ref{sec:method:ppm}) between the substituted and correct triphones. We applied this replacement probability to guide the pseudo-labeling process, replacing triphones in the clean text accordingly.
The $distance$ term ensures that more similar triphones have a higher replacement probability, resulting in a more realistic ASR noise.

We applied the Triphone Pseudo-Noise Labeling method to external text-only corpora (CC-100 \cite{CC1, CC2}), effectively amplifying the P2G noisy training data and enhancing the realism of the ASR noise.

\subsection{Global Phoneme Noise (GPN) generator}
\label{sec:GPN}

Previous methods, such as K-fold and Triphone pseudo-noise labeling, have their respective limitations. K-fold methods suffer from limited sample size, especially for low-resource languages. On the other hand, Triphone pseudo-noise labeling relies on rule-based triphone noise analysis, which may not accurately capture meaningful information.

To overcome these limitations and generate more realistic ASR noise, we adopt a neural network-based approach for generating noisy text from clean phoneme text. Our model architecture utilizes Transformer encoders and Transformer decoders. Figure \ref{fig:arch} illustrates the training and inference process. We train the GPN model using both K-fold noise and Triphone pseudo-labeled noise.
In addition, we enhance the cross-lingual capability of GPN by training a single model using the noise data from all 10 languages.




\section{Experimental Setup}
\label{sec:typestyle}

\subsection{Dataset}

\begin{table}[]
\centering
\resizebox{\columnwidth}{!}{%
\begin{tabular}{c|c|cccc}
Language &
  \begin{tabular}[c]{@{}c@{}}Train set duration\\ (hh:mm)\end{tabular} &
  \begin{tabular}[c]{@{}c@{}}Grapheme \\ vocab size\end{tabular} &
  \begin{tabular}[c]{@{}c@{}}IPA \\ vocab size\end{tabular} &
  \begin{tabular}[c]{@{}c@{}}PPM \\ (K=30, T=2.0)\\ vocab size\end{tabular} &
  \begin{tabular}[c]{@{}c@{}}PPM\\ (K=55, T=0.5)\\ vocab size\end{tabular} \\ \hline \hline
cs & 19:33  & 42 & 47 (0.89) & 23 (1.00) & 38 (1.00) \\
el & 2:04   & 33 & 37 (1.00) & 21 (1.00) & 31 (1.00) \\
es & 397:34 & 80 & 56 (0.91) & 27 (1.00) & 42 (1.00) \\
fi & 2:30   & 26 & 61 (0.67) & 23 (1.00) & 32 (1.00) \\
hu & 10:51  & 34 & 51 (0.88) & 34 (1.00) & 34 (1.00) \\
ka & 6:15   & 32 & 34 (0.85) & 21 (1.00) & 27 (1.00) \\
nl & 38:00  & 36 & 62 (0.90) & 29 (1.00) & 44 (1.00) \\
pl & 23:55  & 35 & 51 (0.74) & 21 (1.00) & 33 (1.00) \\
pt & 21:43  & 38 & 76 (0.78) & 29 (1.00) & 45 (1.00) \\
sl & 1:26   & 25 & 42 (1.00) & 25 (1.00) & 38 (1.00)
\end{tabular}%
}
\caption{Summary of the CommonVoice dataset: Train set duration and vocabulary size for each language, considering various output modeling approaches. The numbers in parentheses indicate the phoneme vocabulary coverage.}
\label{tab:cv}
\end{table}
\textbf{CommonVoice Dataset:} We used the CommonVoice 12.0 dataset, which includes 10 languages with varying available resources. The dataset covers languages ranging from Spanish (397 hours of data) to Slovenian (1 hour and 26 minutes of data). We considered the differences in pronunciation across these languages to ensure a representative set. We resampled all audio files in the CommonVoice dataset to a uniform sampling rate 16,000Hz. We utilized the provided train, dev, and test sets from the CommonVoice 12.0 dataset for our experiments.

\textbf{CC-100 Dataset:} In addition, we utilized the CC-100 web-crawled text corpus \cite{CC1, CC2} for each of the ten languages. The CC-100 dataset contains a significant amount of text with numerous sentences. To effectively manage the dataset, we randomly sampled approximately 5,000,000 sentences per language from the CC-100 corpus. 
The CC-100 data served the primary purpose of P2G pre-training, where the model learns the association between phoneme sequences and grapheme sequences using the text data.

For generating phoneme-level texts, we utilized the Phonimizer tool \cite{Phonemizer} with Espeak-ng backend for both the CommonVoice and CC-100 text datasets.

\subsection{Models and Training details}

\begin{table}[t]
\centering
\resizebox{0.80\columnwidth}{!}{%
\begin{tabular}{r||rrrrr}
\multicolumn{1}{l||}{\begin{tabular}[c]{@{}l@{}}Threshold ($T$)\\ / Pivot ($K$) \end{tabular}} & 0 & 0.5 & 1 & 2 & 3 \\ \hline\hline
0  & 0.87 & 0.87       & 0.87 & 0.87       & 0.87 \\
20 & 0.87 & 0.87       & 0.87 & 1          & 1 \\
30 & 0.87 & 0.89       & 0.89 & \textbf{1} & 1 \\
40 & 0.87 & 0.99       & 0.99 & 1          & 1 \\
55 & 0.87 & \textbf{1} & 1    & 1          & 1 \\
80 & 0.87 & 1          & 1    & 1          & 1
\end{tabular}%
}
\caption{Average phoneme vocabulary coverage across 10 languages. A value of 0 in the Pivot ($K$) and Threshold ($T$) columns indicates that the PPM method was not used (as same as IPA)}
\label{tab:ppm}
\end{table}

\textbf{Phoneme recognition model:} For our phoneme recognition model, we adopted the setup proposed by \cite{Simple} and utilized the XLSR-53 \cite{XLSR-53} weight as a pre-trained model. In order to improve the accuracy of the P2G translation, word delimiter tokens (white space) were added between the phonemes of words. 4-gram language models (LM) were utilized for CTC decoding during the inference and evaluation. Additionally, the LM was trained using phoneme-level text from both the CommonVoice and CC-100 datasets.

Regarding the training procedure, we initially built a base model using English data from the CommonVoice dataset, which served as a general phoneme recognition model. Subsequently, we trained the model using data from all 10 languages to maximize its cross-lingual capability. Finally, we performed fine-tuning for each language using their respective training sets (as indicated in Table \ref{tab:cv}).

\textbf{P2G translator} The P2G translator employed a model architecture consisting of 12 Transformer encoders and 12 Transformer decoders. It was trained using a sentencepiece tokenizer with a vocabulary size 20,000 for each language. The pre-training of the P2G model involved using the CC-100 text data and transcriptions from the CommonVoice train set while fine-tuning incorporated different noisy text train sets, including K-fold, Triphone pseudo-noise labeling, and GPN-generated noise. Separate P2G models were trained for each language and phoneme modeling setup (IPA and PPM). 

\vspace{3mm}

\textbf{Pivot Phoneme Merging (PPM) Setup}
We compare the vocabulary coverage of the PPM method using different parameter settings: the number of pivot phonemes ($K$) and the threshold distance ($T$). Table \ref{tab:ppm} presents the average vocabulary coverage across 10 languages for various values of $K$ and $T$. 
Given that the average IPA vocabulary size across the 10 languages is about 52 (Table \ref{tab:cv}), we consider $K$ values less than 30 to be unrealistic settings. A few pivot phonemes may not sufficiently capture the meaningful differences between tokens in phoneme recognition and P2G models. Moreover, $T$ values larger than 2.0 do not ensure meaningful phoneme merging, as highly distinguishable phonemes may still be merged in such settings.

Table \ref{tab:ppm} indicates that higher threshold distance ($T$) is required for a smaller number of pivots ($K$) in order to cover all phonemes. This result highlights the trade-off between $T$ and $K$ regarding vocabulary coverage. Considering this trade-off, we select two representative settings: (1) $K$=30, $T$=2.0, and (2) $K$=55, $T$=0.5. These settings cover 100\% of the phonemes in their vocabulary and provide distinct differences in $K$ and $T$. In subsequent experiments, these setups will be further evaluated to determine the optimal configuration.



\vspace{-2mm}
\subsection{Global Phoneme Noise (GPN) generator setup}

We employed 6 transformer encoder and decoder architecture with sentencepiece tokenizer as same as P2G translator setup.
To train the GPN model, we utilize noisy data from K-fold (CommonVoice) and the Triphone method (CC-100) (See Section \ref{sec:noise}).

\vspace{-3mm}

\begin{table}[t]
\centering
\resizebox{0.80\columnwidth}{!}{%
\begin{tabular}{c|ccc}
Lang. ID & IPA & \begin{tabular}[c]{@{}c@{}}PPM\\ (K=30, T=2.0)\end{tabular} & \begin{tabular}[c]{@{}c@{}}PPM\\ (K=55, T=0.5)\end{tabular} \\ \hline\hline
cs & \textbf{0.7} & 1.6 & \textbf{0.7} \\
el & \textbf{2.5} & 3.3 & 2.6          \\
es & \textbf{1.3} & 1.7 & 1.6          \\
fi & \textbf{0.4} & 2.8 & 2.3          \\
hu & \textbf{0.7} & 1.2 & 0.8          \\
ka & \textbf{0.9} & 1.1 & \textbf{0.9} \\
nl & \textbf{0.5} & 0.7 & \textbf{0.5} \\
pl & \textbf{0.8} & 1.7 & 1.1 \\
pt & \textbf{0.8} & 1.6 & 1.1          \\
sl & \textbf{0.5} & 0.8 & \textbf{0.5}
\end{tabular}%
}
\caption{Word Error Rate (WER) on the CommonVoice test set using the P2G translation model with correct (clean) input data.}
\label{tab:golden}
\end{table}

\section{Results}

\begin{table}[t]
\centering
\resizebox{0.80\columnwidth}{!}{%
\begin{tabular}{c|ccc}
Lang. ID & IPA & \begin{tabular}[c]{@{}c@{}}PPM \\ (K=30, T=2.0)\end{tabular} & \begin{tabular}[c]{@{}c@{}}PPM \\ (K=55, T=0.5)\end{tabular} \\ \hline \hline
cs & 7.4  & \textbf{5.5}  & 5.9  \\
el & 10.4 & \textbf{5.4}  & 5.7  \\
es & 3.4  & \textbf{2.2}  & 2.5  \\
fi & 11.3 & \textbf{8.2}  & 8.6  \\
hu & 11.2 & \textbf{10.1} & 10.3 \\
ka & 6.6  & \textbf{3.8}  & 4.7  \\
nl & 5.9  & \textbf{5.4}  & 5.5  \\
pl & 4.4  & \textbf{3.1}  & 3.5  \\
pt & 6.5  & \textbf{5.5}  & 5.6  \\
sl & 9.6  & \textbf{8.2}  & 8.4 
\end{tabular}%
}
\caption{Phoneme Error Rate (PER) in \% on the CommonVoice test set for various phoneme modeling approaches. Each model incorporates a 4-gram language model (LM) during the CTC decoding process.}
\label{tab:PER}
\end{table}

\subsection{Phoneme Recognition}


Table \ref{tab:PER} presents the Phoneme Error Rate (PER) for both IPA and PPM modeling. The results demonstrate significant improvements in PER for all 10 languages with PPM models compared to the IPA model, which serves as the baseline. Notably, both PPM models, (K=30, T=2.0) and (K=55, T=0.5), achieve 100\% phoneme vocabulary coverage (Table \ref{tab:ppm}), but the (K=30, T=2.0) model outperforms the (K=55, T=0.5) model due to its more compact vocabulary size and greater vocabulary sharing. It is worth mentioning that PPM models have an advantage in phoneme recognition as their output units are a simplified version of IPA. Similarly, within the PPM models, (K=30, T=2.0) is easier to recognize than (K=55, T=0.5).
\vspace{-4mm}

\subsection{P2G translation}
\vspace{-1mm}
Table \ref{tab:best} presents the results of P2G translation for both IPA and PPM models. To establish the superiority of our methods over traditional grapheme-level modeling, we conducted experiments using a grapheme-level model with the same experimental setup as the other models, utilizing the XLSR-53 weight as the pre-trained model and CTC head on top. The P2G translators were fine-tuned to maximize their performance using K-fold noise, Triphone pseudo-noise, and GPN-generated noise. 

Among the different models, PPM (K=55, T=0.5) with P2G achieved the best WER in all 10 languages, exhibiting a relative reduction in WER of approximately 19\% compared to the Grapheme baseline and around 14\% compared to the IPA model.
Our method demonstrates superior performance in low-resource languages and with sufficient resources, such as Spanish (es).
Although PPM with (K=30, T=2.0) showed the best phoneme recognition results in Table \ref{tab:PER}, it sacrificed too much representation of IPA tokens, leading to suboptimal P2G results. Further PPM and P2G relationship analysis is conducted in Section \ref{sec:effect-of-ppm}.

\begin{table*}[t]
\centering
\resizebox{0.80\textwidth}{!}{%
\begin{tabular}{lllllllllllr}
Lang. ID  & cs   & el   & es  & fi   & hu   & ka   & nl   & pl   & pt   & sl   & \multicolumn{1}{l}{avg.} \\ \hline
\multicolumn{12}{l}{Model}                                                                                \\ \hline\hline
Grapheme  & 31.6 & 35.6 & 8.5 & 42.2 & 39.9 & 25.4 & 14.7 & 19.7 & 18.1 & 34.9 & 27.1                     \\ \hline
IPA + P2G & 26.8 & 29.0 & 8.4 & 40.2 & 39.6 & 27.0 & 15.6 & 17.6 & 17.2 & 33.6 & 25.5                     \\ \hline
\begin{tabular}[c]{@{}l@{}}PPM + P2G\\ (K=30, T=2.0)\end{tabular} &
  24.3 &
  28.8 &
  6.8 &
  38.2 &
  38.8 &
  22.1 &
  14.4 &
  14.2 &
  16.4 &
  29.1 &
  23.3 \\ \hline
\begin{tabular}[c]{@{}l@{}}PPM + P2G\\ (K=55, T=0.5)\end{tabular} &
  \textbf{21.8} &
  \textbf{26.0} &
  \textbf{6.5} &
  \textbf{37.5} &
  \textbf{36.3} &
  \textbf{21.2} &
  \textbf{13.2} &
  \textbf{13.6} &
  \textbf{15.3} &
  \textbf{28.5} &
  \textbf{22.0}
\end{tabular}%
}
\caption{Results of Phoneme Recognition + P2G Translation: Word Error Rate (WER) in \% on the CommonVoice test set. All results are based on Grapheme-level text.}
\label{tab:best}
\end{table*}

\vspace{-3mm}
\section{Analysis}

\subsection{Effect of PPM on Phoneme recognition and P2G translation}
\label{sec:effect-of-ppm}

Table \ref{tab:best} reveals that in certain languages, the IPA+P2G architecture exhibits higher Word Error Rate (WER) compared to the Grapheme model. This can be attributed to the challenge of achieving low PER in the IPA model (Table \ref{tab:PER}), which results in error propagation during the P2G process. However, the PPM method proves effective in surpassing both the grapheme and IPA baselines, indicating that appropriate parameter settings in PPM can optimize both phoneme recognition and P2G translation within a two-pass ASR system.

In order to assess the impact of PPM on P2G translation directly, we conducted experiments using clean text input. Table \ref{tab:golden} demonstrates that IPA achieves the best result for P2G translation since it does not lose any representatives caused by phoneme merging in PPM. The increase in homophones during PPM can result in higher WER in P2G translation. While PPM with (K=30, T=2.0) leads to degraded WER in all languages, PPM with (K=55, T=0.5) shows more reliable results. 
This result indicates that PPM with (K=55, T=0.5) effectively can keep each language IPA representative because of the minimum distance threshold (T=0.5).  

\subsection{Effect of GPN settings in P2G translation}
\label{sec:effect-of-GPN}


In order to evaluate the effect of GPN on the P2G translation process, we conducted experiments on two languages: Portuguese and Finnish (Table \ref{tab:GPN}). 
By comparing P2G models trained with different noise data from various noise generation methods, we observed that noise generated by K-fold significantly improves P2G translation performance. Triphone pseudo-noise labeling provides additional improvements, and incorporating GPN-generated noise further enhances the results. Notably, using GPN with multiple languages yields even better performance.

These findings emphasize the efficacy of incorporating noise-aware P2G training, particularly in low-resource language scenarios like Finnish, where the higher PER (refer to Table \ref{tab:PER}) has a substantial impact on the quality of P2G result (WER). The overall results underscore the positive influence of integrating the GPN approach into the P2G translation process, demonstrating its potential to enhance cross-lingual accuracy in Two-Pass ASR systems.


\begin{table}[h!]
\resizebox{0.85\columnwidth}{!}{%
\begin{tabular}{lll}
Lang. ID (train set duration)                                                           & pt (21:43)    & fi (2:30)     \\ \hline \hline
w/o noise                                                                               & 16.9          & 42.5          \\ \hline
+ K-fold                                                                                 & 16.1          & 38.9          \\ \hline
\begin{tabular}[c]{@{}l@{}}+ K-fold\\ + Triphone\end{tabular}                           & 15.8          & 38.6          \\ \hline
\begin{tabular}[c]{@{}l@{}}+ K-fold\\ + Triphone\\ + GPN (single language)\end{tabular} & 15.6          & 37.8          \\ \hline
\begin{tabular}[c]{@{}l@{}}+ K-fold\\ + Triphone\\ + GPN (10 language)\end{tabular}     & \textbf{15.3} & \textbf{37.5}
\end{tabular}%
}
\caption{P2G results (WER) on CommonVoice test set, evaluating the impact of noise train set on P2G. PPM (K=55, T=0.5) used for languages pt and fi.}
\label{tab:GPN}
\end{table}
\vspace{-3mm}




\vspace{-3mm}
\section{Conclusion}

In conclusion, our integrated approach combining Phoneme Pivot Method (PPM) and Global Phoneme Noise (GPN) generator improves Two-Pass automatic speech recognition (ASR) systems. PPM enhances phoneme recognition, while GPN enhances P2G translation, resulting in an approximate 19\% relative reduction in WER compared to the baseline. The incorporation of GPN-generated noise improves the training of P2G models, particularly benefiting low-resource languages. This approach enhances cross-lingual ASR performance, optimizes vocabulary sharing, and increases the robustness and adaptability of Two-Pass ASR systems. Our research contributes to the advancement of cross-lingual Two-Pass ASR and the optimization of noise-aware training methods.

\section{Acknowledgement}
\footnotesize{This work was supported by the Technology Innovation Program(20015007, Development of Digital Therapeutics of Cognitive Behavioral Therapy for treating Panic Disorder) funded By the Ministry of Trade, Industry \& Energy(MOTIE, Korea)", 
\noindent This research is supported by Culture Technology R\&D Program through the Korea Creative Content Agency funded by Ministry of Culture, Sports and Tourism (Development of contents meta-verse based on XR and AI, R2021040136)}

\normalsize
\newpage
\bibliographystyle{IEEEbib}
\bibliography{strings,refs}

\begin{thebibliography}{10}

\bibitem{JUST}
Junwen Bai, Bo~Li, Yu~Zhang, Ankur Bapna, Nikhil Siddhartha, Khe~Chai Sim, and Tara~N. Sainath,
\newblock ``Joint unsupervised and supervised training for multilingual asr,''
\newblock in {\em International Conference on Acoustics, Speech and Signal Processing (ICASSP)}, 2022, pp. 6402--–6406.

\bibitem{M-CTC}
Loren Lugosch, Tatiana Likhomanenko, Gabriel Synnaeve, and Ronan Collobert,
\newblock ``Pseudo-labeling for massively multilingual speech recognition,''
\newblock in {\em ICASSP 2022 - 2022 IEEE International Conference on Acoustics, Speech and Signal Processing (ICASSP)}, 2022, pp. 7687--7691.

\bibitem{multilingual}
Shubham Toshniwal, Tara~N. Sainath, Ron Weiss, Bo~Li, Pedro Moreno, Eugene Weinsten, and Kanishka Rao,
\newblock ``Multilingual speech recognition with a single end-to-end model,''
\newblock 2018.

\bibitem{Simple}
Qiantong Xu, Alexei Baevski, and Michael Auli,
\newblock ``Simple and effective zero-shot cross-lingual phoneme recognition,''
\newblock in {\em Interspeech}, 2022.

\bibitem{UniSpeech}
Chengyi Wang, Yu~Wu, Yao Qian, Kenichi Kumatani, Shujie Liu, Furu Wei, Michael Zeng, and Xuedong Huang,
\newblock ``Unispeech: Unified speech representation learning with labeled and unlabeled data,''
\newblock in {\em Proceedings of the 38th International Conference on Machine Learning, {ICML} 2021, 18-24 July 2021, Virtual Event}, Marina Meila and Tong Zhang, Eds. 2021, vol. 139 of {\em Proceedings of Machine Learning Research}, pp. 10937--10947, {PMLR}.

\bibitem{Globalphone}
Tanja Schultz,
\newblock ``{Globalphone: a multilingual speech and text database developed at karlsruhe university},''
\newblock in {\em Proc. 7th International Conference on Spoken Language Processing (ICSLP 2002)}, 2002, pp. 345--348.

\bibitem{IPA}
I.P.Association,
\newblock ``Handbook of the international phonetic association: A guide to the use of the international phonetic alphabet,''
\newblock in {\em Cambridge University Press}, 1999.

\bibitem{survey}
Laurent Besacier, Etienne Barnard, Alexey Karpov, and Tanja Schultz,
\newblock ``Automatic speech recognition for under-resourced languages: A survey,''
\newblock {\em Speech Communication}, vol. 56, pp. 85--100, 2014.

\bibitem{wav2vec2}
Alexei Baevski, Henry Zhou, Abdelrahman Mohamed, and Michael Auli,
\newblock ``Wav2vec 2.0: A framework for self-supervised learning of speech representations,''
\newblock in {\em Proceedings of the 34th International Conference on Neural Information Processing Systems}, Red Hook, NY, USA, 2020, NIPS'20, Curran Associates Inc.

\bibitem{allosaurus}
Xinjian Li, Siddharth Dalmia, Juncheng Li, Matthew Lee, Patrick Littell, Jiali Yao, Antonios Anastasopoulos, David~R Mortensen, Graham Neubig, Alan~W Black, and Metze Florian,
\newblock ``Universal phone recognition with a multilingual allophone system,''
\newblock in {\em ICASSP 2020-2020 IEEE International Conference on Acoustics, Speech and Signal Processing (ICASSP)}. IEEE, 2020, pp. 8249--8253.

\bibitem{sc-BERT}
Oleksii Hrinchuk, Mariya Popova, and Boris Ginsburg,
\newblock ``Correction of automatic speech recognition with transformer sequence-to-sequence model,''
\newblock in {\em ICASSP 2020 - 2020 IEEE International Conference on Acoustics, Speech and Signal Processing (ICASSP)}, 2020, pp. 7074--7078.

\bibitem{sc-kfold}
Shuai Zhang, Jiangyan Yi, Zhengkun Tian, Ye~Bai, Jianhua Tao, Xuefei Liu, and Zhengqi Wen,
\newblock ``{End-to-End Spelling Correction Conditioned on Acoustic Feature for Code-Switching Speech Recognition},''
\newblock in {\em Proc. Interspeech 2021}, 2021, pp. 266--270.

\bibitem{sc-rnnt}
Jinyu Li, Rui Zhao, Zhong Meng, Yanqing Liu, Wenning Wei, Sarangarajan Parthasarathy, Vadim Mazalov, Zhenghao Wang, Lei He, Sheng Zhao, and Yifan Gong,
\newblock ``Developing rnn-t models surpassing high-performance hybrid models with customization capability,'' 2020.

\bibitem{sc-TTS}
Jinxi Guo, Tara Sainath, and Ron Weiss,
\newblock ``A spelling correction model for end-to-end speech recognition,''
\newblock 05 2019.

\bibitem{first}
Peter Pol{\'a}k, Sangeet Sagar, Dominik Mach{\'a}{\v{c}}ek, and Ond{\v{r}}ej Bojar,
\newblock ``{CUNI} neural {ASR} with phoneme-level intermediate step for{\textasciitilde}{N}on-{N}ative{\textasciitilde}{SLT} at {IWSLT} 2020,''
\newblock in {\em Proceedings of the 17th International Conference on Spoken Language Translation}, Online, July 2020, pp. 191--199, Association for Computational Linguistics.

\bibitem{tranUSR}
Hongfei Xue, Qijie Shao, Peikun Chen, Pengcheng Guo, Lei Xie, and Jie Liu,
\newblock ``Tranusr: Phoneme-to-word transcoder based unified speech representation learning for cross-lingual speech recognition,'' 2023.

\bibitem{panphon}
David~R. Mortensen, Patrick Littell, Akash Bharadwaj, Kartik Goyal, Chris Dyer, and Lori~S. Levin,
\newblock ``Panphon: {A} resource for mapping {IPA} segments to articulatory feature vectors,''
\newblock in {\em Proceedings of {COLING} 2016, the 26th International Conference on Computational Linguistics: Technical Papers}. 2016, pp. 3475--3484, {ACL}.

\bibitem{CC1}
Guillaume Wenzek, Marie-Anne Lachaux, Alexis Conneau, Vishrav Chaudhary, Francisco Guzm{\'a}n, Armand Joulin, and Edouard Grave,
\newblock ``{CCN}et: Extracting high quality monolingual datasets from web crawl data,''
\newblock in {\em Proceedings of the Twelfth Language Resources and Evaluation Conference}, Marseille, France, May 2020, pp. 4003--4012, European Language Resources Association.

\bibitem{CC2}
Alexis Conneau, Kartikay Khandelwal, Naman Goyal, Vishrav Chaudhary, Guillaume Wenzek, Francisco Guzm{\'a}n, Edouard Grave, Myle Ott, Luke Zettlemoyer, and Veselin Stoyanov,
\newblock ``Unsupervised cross-lingual representation learning at scale,''
\newblock in {\em Proceedings of the 58th Annual Meeting of the Association for Computational Linguistics}, Online, July 2020, pp. 8440--8451, Association for Computational Linguistics.

\bibitem{Phonemizer}
Mathieu Bernard and Hadrien Titeux,
\newblock ``Phonemizer: Text to phones transcription for multiple languages in python,''
\newblock {\em Journal of Open Source Software}, vol. 6, no. 68, pp. 3958, 2021.

\bibitem{XLSR-53}
Alexis Conneau, Alexei Baevski, Ronan Collobert, Abdelrahman Mohamed, and Michael Auli,
\newblock ``{Unsupervised Cross-Lingual Representation Learning for Speech Recognition},''
\newblock in {\em Proc. Interspeech 2021}, 2021, pp. 2426--2430.

\end{thebibliography}

\end{document}